\documentclass[lettersize,journal]{IEEEtran}
\usepackage{amsmath,amsfonts}
\usepackage{algorithmic}
\usepackage{algorithm}
\usepackage{array}
\usepackage[caption=false,font=normalsize,labelfont=sf,textfont=sf]{subfig}
\usepackage{textcomp}
\usepackage{stfloats}
\usepackage{url}
\usepackage{verbatim}
\usepackage{graphicx}
\usepackage{cite}
\usepackage{hyperref}

\usepackage{amsthm}
\usepackage{booktabs}
\usepackage[switch]{lineno}
\usepackage{amssymb}
\usepackage{multirow}

\newtheorem{definition}{Definition}

\begin{document}

\title{ULFine: Unbiased Lightweight Fine-tuning for Foundation-Model-Assisted Long-Tailed Semi-Supervised Learning}

\author{Enhao~Zhang,
        Chaohua~Li,
        Chuanxing~Geng,    
        and~Songcan~Chen,~\IEEEmembership{Senior~Member,~IEEE}
\IEEEcompsocitemizethanks{\IEEEcompsocthanksitem E. Zhang, C. Geng, C. Li, and S. Chen are with MIIT Key Laboratory of Pattern Analysis and Machine Intelligence, China
and College of Computer Science and Technology, Nanjing University of Aeronautics and Astronautics (NUAA), Nanjing 211106, China\protect\\
E-mail:\{zhangeh, chaohuali, gengchuanxing, s.chen\}@nuaa.edu.cn.}

\thanks{Corresponding authors: Songcan Chen}
}

\markboth{Journal of \LaTeX\ Class Files,~Vol.~14, No.~8, August~2021}%
{Shell \MakeLowercase{\textit{et al.}}: A Sample Article Using IEEEtran.cls for IEEE Journals}


\maketitle

\begin{abstract}
Based on the success of large-scale visual foundation models like CLIP in various downstream tasks, this paper initially attempts to explore their impact on Long-Tailed Semi-Supervised Learning (LTSSL) by employing the foundation model with three strategies: Linear Probing (LP), Lightweight Fine-Tuning (LFT) and Full Fine-Tuning (FFT). Our analysis presents the following insights: i) Compared to LTSSL algorithms trained from scratch, FFT results in a decline in model performance, whereas LP and LFT, although boosting overall model performance, exhibit negligible benefits to tail classes.
ii) LP produces numerous false pseudo-labels due to \textit{underlearned} training data, while LFT can reduce the number of these false labels but becomes overconfident about them owing to \textit{biased fitting} training data.
This exacerbates the pseudo-labeled and classifier biases inherent in LTSSL, limiting performance improvement in the tail classes.
With these insights, we propose a Unbiased Lightweight Fine-tuning strategy, \textbf{ULFine}, which mitigates the overconfidence via confidence-aware adaptive fitting of textual prototypes and counteracts the pseudo-labeled and classifier biases via complementary fusion of dual logits.
Extensive experiments demonstrate that ULFine markedly decreases training costs by over ten times and substantially increases prediction accuracies compared to state-of-the-art methods.
\end{abstract}

\begin{IEEEkeywords}
Long-tailed semi-supervised learning, foundation model,  lightweight fine-tuning, pseudo labels.
\end{IEEEkeywords}

\section{Introduction}

\begin{figure*}[t]
  \centering
   \includegraphics[width=0.9\linewidth]{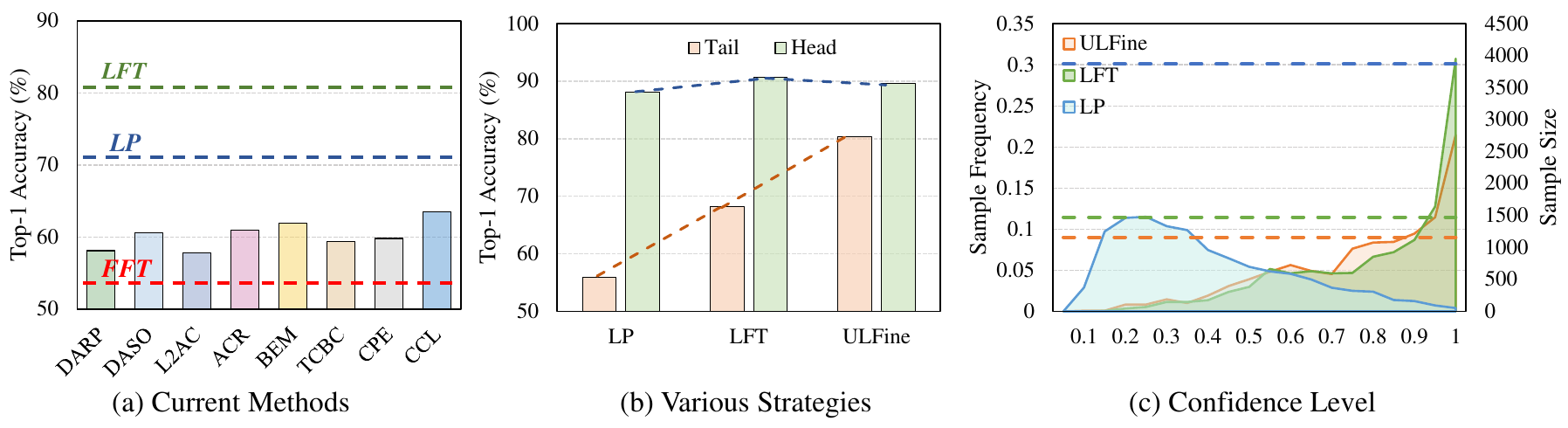}
   \caption{On the CIFAR100-LT dataset, (a): Comparison of top-1 accuracy of Linear Probing (LP), Lightweight Fine-Tuning (LFT), and Full Fine-Tuning (FFT) with existing LTSSL methods. (b) Comparison of top-1 accuracy of head and tail classes of different strategies. (c) The vertical axes (left and right) indicate the confidence level (area plot) and the sample size (dashed line) of false pseudo-labeling.}
   \label{fig1}
\end{figure*}

\IEEEPARstart{S}{emi}-supervised learning (SSL) represents a key strategy for improving the generalizability of deep neural networks by utilizing limited labeled samples and massive unlabeled samples \cite{xu2021dash,fan2022cossl}. 
Typical SSL algorithms employ consistency constraints to generate pseudo-labels for unlabeled samples and select reliable ones to participate in model training \cite{wang2022freematch,wu2024allmatch,sohn2020fixmatch}.
These algorithms generally assume that labeled and unlabeled samples obey a uniform distribution. However, subject to power-law distributions, real-world datasets tend to exhibit long-tailed distributions \cite{zhang2024dynamic,weilearning,liu2019large}. This discrepancy inevitably leads to biased pseudo-labels and classifiers, exacerbating the class imbalance during training and ultimately hindering model performance \cite{zhou2024continuous}.

In response to these problems, long-tailed semi-supervised learning (LTSSL) has received widespread attention in recent years. It usually assumes that labeled samples obey long-tailed distributions yet the distribution of unlabeled samples is unknown and potentially mismatched with those of labeled samples.
Current LTSSL methods typically cope with imbalance dilemma by leveraging techniques such as logit adjustment \cite{wei2023towards,ma2024three,zhou2024continuous}, distribution alignment \cite{kim2020distribution,wei2021crest} and adaptive threshold \cite{guo2022class,lai2022smoothed}.
Despite substantial progress, they generally adopt a scratch training strategy that constrains the model's generalizability, thereby failing to effectively mitigate the intractable pseudo-labeled bias and classifier bias.
Instead of training neural networks from scratch, recent studies have shown that applying pre-trained foundation models like CLIP \cite{radford2021learning} to various downstream tasks demonstrates impressive generalization capabilities, including supervised Long-Tailed Recognition (LTR) \cite{wenmakes,shi2024long}, out-of-distribution detection \cite{li2024learning}, and few-shot learning \cite{miyai2024locoop}.
However, the potential of the foundation model to enhance LTSSL performance remains unexplored.

To unleash their potential in LTSSL, we pilotly explore the \textit{global overall performance} impact of employing the foundation model with various strategies, \textit{i.e.}, Linear Probing (LP), Lightweight Fine-Tuning (LFT), and Full Fine-Tuning (FFT). From the results presented in Fig. \ref{fig1} (a), we can observe: 
1) When LP is employed, where the foundational model (CLIP) is frozen and only the classifier is trained, which outperforms all current methods.
2) When the well-respected LFT in supervised LTR is adopted, where only a small portion of the parameters are updated, the model's performance obtains further improvements.
3) When FFT is implemented, where the entire neural network is updated, the model's performance is significantly degraded.
The issue arises because FFT destroys intra-class distance distribution, resulting in inconsistent class-conditional probabilities for tail classes in training and test sets \cite{shi2024long}. 

Furthermore, we explore the \textit{local grouping performance} illustrated in Fig. \ref{fig1} (b). 
We discover that LP and LFT exhibit an excessive focus on the head classes of labeled samples (hereafter referred to as head classes), while the tail classes, which deserve more attention, are almost neglected, regardless of the number of unlabeled samples, termed as “\textit{minority bottleneck}”.
Moreover, we reveal that LP, limited to training classifiers alone, produces numerous false pseudo-labels due to underlearning of the training data. 
Although LFT can reduce the number of false pseudo-labels, it exhibits undesirable overconfidence in them owing to the biased fitting of the training data, termed as "\textit{majority overconfidence}", as shown in Fig. \ref{fig1} (c) (detailed analysis in Sec. \ref{analysis}).
In the semi-supervised training paradigm, these samples with overconfident false pseudo-labels are hardly filtered by the masker, exacerbating biases in pseudo-labels and classifiers during training, and ultimately hindering improvements in tail class performance.

To overcome the above problems, we propose a simple and effective \textbf{U}nbiased \textbf{L}ightweight \textbf{F}ine-tuning strategy, \textbf{ULFine}, which consists of two core components, Prototype Adaptive Fitting (PAF) and Dual Logit Fusion (DLF). 
Specifically, PAF adaptively draws on visual prototype knowledge by confidence-aware pseudo-labeling distributions to encourage the foundation model to fit downstream imbalanced classification tasks unbiasedly. 
Meanwhile, it introduces orthogonality constraints to refine both visual and textual prototypes to avoid overconfidence in head classes.
On the other hand, inspired by the complementary nature of the pseudo-labels obtained from the similarity and linear classifiers in \cite{oh2022daso}, the DLF is designed to seamlessly align and fuse logits from the unbiased textual prototypes and the linear probing, respectively, to obtain comprehensive knowledge against unknown complex distributions. Such enhanced logits not only can facilitate the generation of unbiased pseudo-labels but also mitigate classifier bias.
As shown in Fig. \ref{fig1}, ULFine not only maintains the performance of head classes but also achieves notable improvements in tail classes, while significantly reducing the number of false pseudo-labels and alleviating their overconfidence problem.
In summary, our main contributions are as follows.

\begin{itemize}
    \item  We attempt to explore the impact of the foundation model on LTSSL and discover that FFT degrades the global overall performance, while LP and LFT make a positive contribution.
    \item Our analysis reveals that employing LP suffers from the “minority bottleneck” issue. Although the introduction of LFT can alleviate this problem slightly, it encounters the “majority overconfidence” dilemma.
    \item We propose an unbiased ULFine strategy that not only alleviates the “minority bottleneck” and “majority overconfidence”, but also mitigates the pseudo-labeling and classifier biases by inventing PAF and DLF.
    \item On multiple benchmark datasets, we validate that our ULFine not only significantly decreases the training cost by over ten times but also drastically increases the model performance compared to state-of-the-art methods.
\end{itemize}

\section{Related Works}

\begin{figure*}[t]
  \centering
   \includegraphics[width=0.95\linewidth]{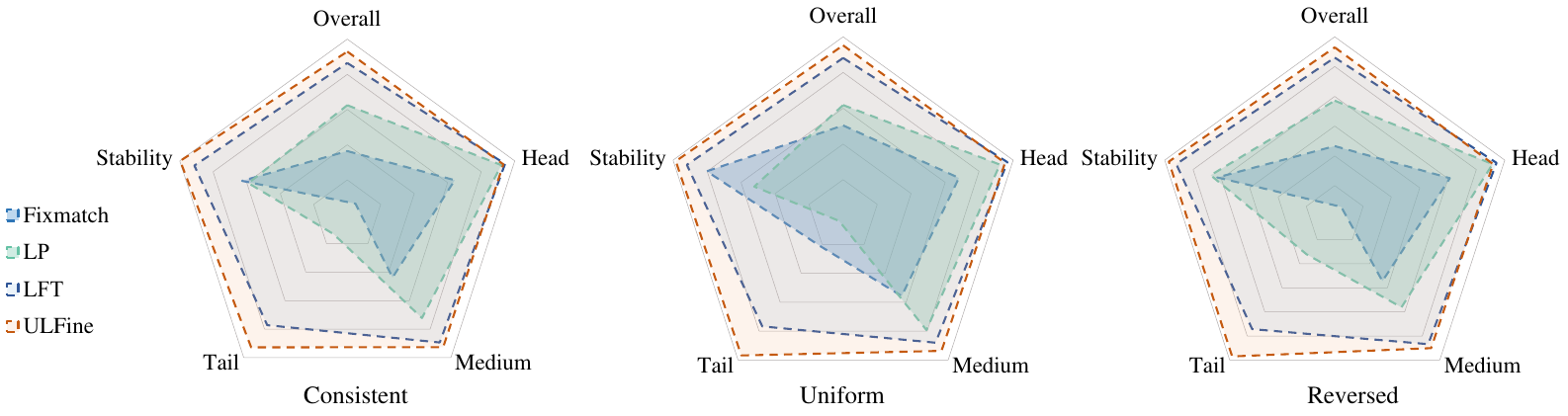}
   \caption{
   Performance comparison of various methods on the CIFAR10-LT with $N_1$=500, $M_1$=4000, and $\gamma_l$=100. “Consistent”, “Uniform” and “Reversed” correspond to scenarios where the imbalance rate $\gamma_u$ is “100", “1", and “1/100" for the unlabeled dataset, respectively.}
   \label{fig2}
\end{figure*}

\subsection{Long-Tailed Semi-Supervised Learning}
In recent years, long-tailed semi-supervised learning has received widespread attention due to practical applications in real-world scenarios. For instance, DARP \cite{kim2020distribution} utilizes distributional alignment techniques to correct biased pseudo-labels thereby mitigating model bias. 
However, these initial methods typically assume that the distribution of unlabeled samples is known and aligned with the labeled ones. When confronted with more realistic scenarios, the model's performance suffers a significant degradation \cite{feng2024bacon,lee2021abc}. 
To address this challenge, subsequent LTSSL methods typically integrate the estimated pseudo-labeled sample distribution into the rebalancing strategy to mitigate the complex imbalance problem. For example, \cite{wei2023towards} corrects the classifier to estimate the true class distribution of unlabeled samples by introducing an adaptive consistency regularizer. Although these approaches have made some progress, they do not significantly improve model performance compared to the balanced scenario. 
To this end, this paper introduces the foundation model with impressive generalisability to LTSSL.
This is not trivial, as we discover the annoying "minority bottleneck" and "majority overconfidence" phenomena with existing strategies to employ the foundation model. 
This paper proposes the unbiased ULFine strategy, which can simultaneously solve the above dilemmas and significantly improve model performance.

\subsection{Vision-Language Foundation Models}
Vision-language foundation models pre-trained with contrastive learning strategies have achieved remarkable success in image-text representation learning. For instance, CLIP introduces a large-scale natural language-supervised approach for open-vocabulary zero-shot image classification. Similarly, ALIGN \cite{jia2021scaling} aligns visual and linguistic representations in a shared latent space, demonstrating robust performance even with noisy image-text pairs. SLIP \cite{mu2022slip} further combines CLIP’s loss function with self-supervised objectives during pre-training. Meanwhile, CoCa \cite{yu2022coca} unifies contrastive loss and captioning loss to pre-train image-text foundation models, thereby inheriting the advantages of both contrastive methods (e.g., CLIP) and generative approaches (e.g., SimVLM \cite{wang2021simvlm}), leading to significant improvements across downstream tasks.
In this paper, we pioneer the application of pre-trained foundation models to LTSSL tasks and systematically investigate the impact of vision-language pre-trained models (exemplified by CLIP) under different fine-tuning strategies.

\subsection{Long-Tail Learning with Foundation Model}

In supervised long-tailed recognition, there has been some work on mitigating the imbalance problem with the help of foundation models to improve model performance \cite{he2023uniformly,tian2022vl,zhao2024ltgc}. For example, BALLAD \cite{ma2021simple} is trained to perform long-tailed recognition by continuing to train and then fixing the visual-language model. LPT \cite{dong2022lpt} motivates pre-trained models to adapt to long-tailed data by dividing prompts into shared prompt and group-specific prompts. Recently, \cite{shi2024long} disclosed that heavy fine-tuning may even lead to non-negligible performance degradation on tail classes, while lightweight fine-tuning is more effective. 
However, these methods only focus on imbalanced learning in the supervised scenario and may fail when faced with more challenging semi-supervised scenarios. This is because confronted with vast quantities of unlabeled samples, the foundation model struggles to appropriately adapt to the downstream tasks, leading to underlearning or biased fitting of the training data.
Consequently, we propose an unbiased lightweight fine-tuning strategy that adaptively fits long-tailed semi-supervised samples, which in turn achieves unbiased pseudo-labeled distributions and classifiers.

\section{Problem Setup and Analysis}

\subsection{Problem Setup}
In LTSSL, the usual setup is a training set with labeled set $\mathbf{\mathcal{X}} =\left \{ \left (\mathbf{x}_n, y_n \right )  \right \} _{n=1}^{N}$ and unlabeled set $\mathbf{\mathcal{U}}=\left\{\mathbf{u}_{m}\right\}_{m=1}^M$, where $y_n\in [C]$ denotes the ground-truth, $N$ and $M$ denote the number of labeled and unlabeled samples. 
The number of labeled and unlabeled samples in the  $c$-th class is defined
$N_c$ and $M_c$, $N =  {\textstyle \sum_{c=1}^{C}} N_c$ and $M =  {\textstyle \sum_{c=1}^{C}} M_c$, where $M_c$ is unknown. Without loss of generality, we assume that the $C$ classes are sorted in descending order, \textit{i.e.}, $N_1\ge N_2\ge \cdots \ge N_C$ and all subsequent features and prototypes are $\ell_2$-normalized.
We denote the imbalance rates of the labeled and unlabeled samples as $\gamma_l=N_1/N_C$ and $\gamma_u=max\left \{ M_c \right \}/min\left \{ M_c \right \}$, respectively.

Following the usual LTSSL studies, this paper is based on a typical SSL framework, \textit{i.e.}, FixMatch \cite{sohn2020fixmatch}. Specifically, using the standard cross-entropy loss $\mathcal{H}$ can be formalized as follows,
\begin{eqnarray}
\begin{aligned}
\mathcal{L}_{F} & =\frac{1}{B_l} \sum_{i=1}^{B_l} \mathcal{H}\left(y_i,p(y|\mathbf{x}_i)\right) \\
& +\frac{1}{B_u} \sum_{j=1}^{B_u} \mathcal{M} \cdot \mathcal{H}\left( q_j,p(y|\mathcal{A}_s({\mathbf{u}}_j) \right),
\label{eq1}
\end{aligned}
\end{eqnarray}
where $p(y|\mathbf{x}_i)=Softmax(f(\mathbf{x}_i); \theta )$ denotes the posterior
probability of $\mathbf{x}_i$ being classified into class $y$.
$\mathcal{M} = \mathbb{I}\left(max\left(p(y|\mathcal{A}_w\left(\mathbf{u}_j\right))\right)>\tau\right)$ is the mask to filter low-confidence pseudo-labels with
a threshold $\tau$ and $\mathbb{I}$ is the indicator function, and  $q_j=argmax_k(q_{jk}))$ is the pseudo-label of $\mathbf{u}_j$.
$B_l$ and $B_u$ represent the number of labeled and unlabeled samples in a mini-batch, respectively.
$\mathcal{A}_s$ and $\mathcal{A}_w$ correspond to\textit{ strong}  and\textit{ weak} augmentation, respectively.

\subsection{Problem Analysis}
\label{analysis}

\begin{figure*}[t]
  \centering
   \includegraphics[width=0.95\linewidth]{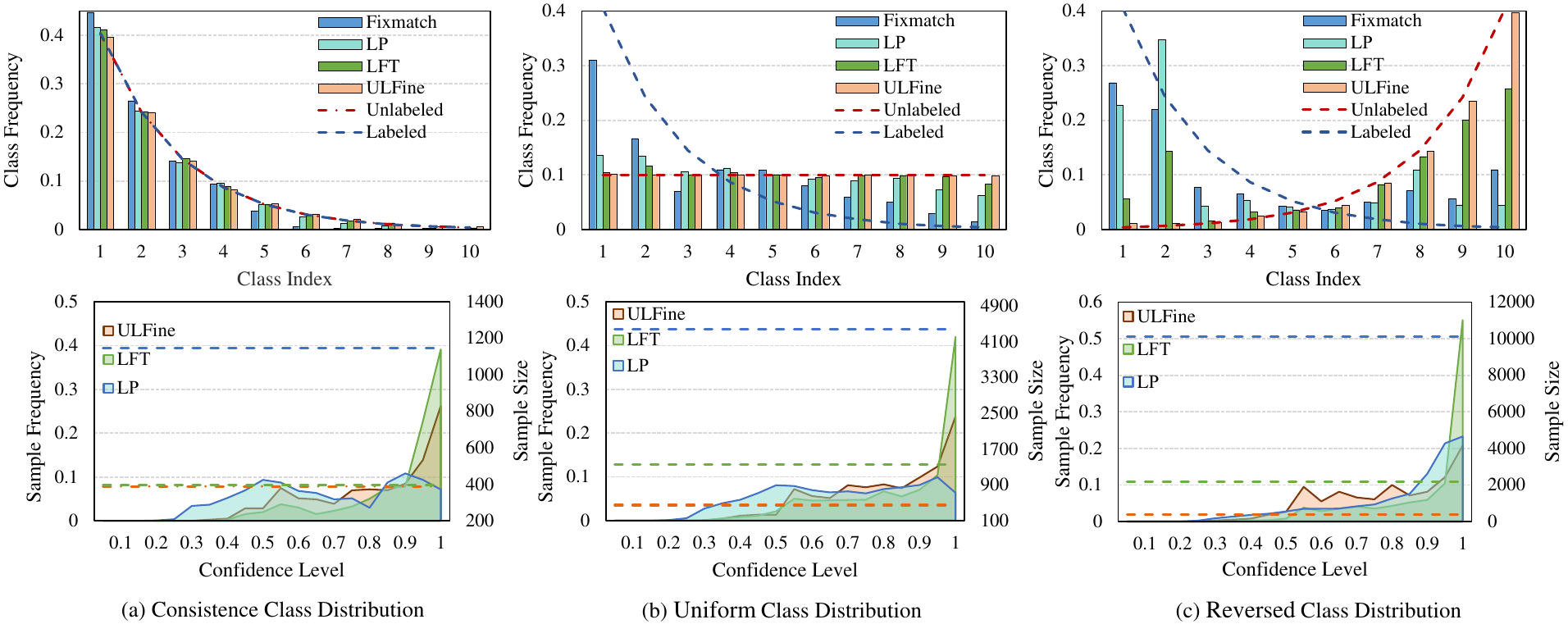}
   \caption{Statistics of relevant results on CIFAR10-LT. \textit{Top:} Distribution of pseudo-labeled samples obtained using different strategies (bar graph). The dotted lines indicate true distributions. \textit{Bottom}: 
   The vertical axes (left and right) indicate the confidence level (area plot) and the sample size (dashed line) of false pseudo-labeling.}
   \label{fig3}
\end{figure*}

\subsubsection{Minority Bottleneck}
To better delineate the local performance of the model, we mimic the supervised LTR to group the classes based on the intra-class sample size of the labeled set. Specifically, “Head” and “Tail” refer to the classes where the intra-class sample size of the labeled set is “$\ge100$ ” and “$\le 20$”, respectively, and the remaining classes are labeled as “Medium". In addition, to characterize the classifier's balance degree, we define the \textit{classification stability}, as outlined in Definition \ref{definition1}.
\begin{definition}
\label{definition1}
     (Classification Stability.)  For a given dataset $\mathbf{\mathcal{D}} =\left \{ \left (\mathbf{x}_i, y_i \right )  \right \} _{i=1}^{N}$, where the probability of sample $\mathbf{x}_i$ being classified correctly is  $p_i=p_{\mathbf{x}|y_i}\left ( y_i=argmax_y p\left ( y|\mathbf{x}_i \right )  \right ) $, then its corresponding classification stability can be formally defined as, 
     \begin{eqnarray}
     S=1-\sqrt{\frac{1}{N}\sum_{i=1}^{N} \left ( p_i-\frac{1}{N}\sum_{j=1}^{N}p_j   \right )^2 }.
     \end{eqnarray}
\end{definition}
Classification stability reflects the model's capacity to classify all samples, with higher values of $S$ indicating a more balanced classifier and vice versa. 
Utilizing the above definitions, we examine the \textit{local grouping performance} of the model on the CIFAR10-LT dataset, with the experimental outcomes depicted in Fig. \ref{fig2}.
Notably, LP significantly enhances the performance of head classes, while the tail classes remain entangled with Fixmatch, regardless of the pseudo-labeled sample distribution being consistent, uniform, or reversed. We define this phenomenon as the “\textit{minority bottleneck}". 
Furthermore, one can notice that even introducing LFT, which has claimed to mitigate the effects of imbalance, only gains limited improvements in both tail classes and classification stability.
To delve into the underlying causes, we statistically analyze the predictive distribution of pseudo-labeling on CIFAR10-LT. Based on the results in Fig. \ref{fig3} (Top), we conclude that the primary reasons are as follows.

\begin{itemize}
    \item When the class distributions are consistent, LP and LFT can obtain a relatively satisfactory pseudo-labeling distribution than Fixmatch. However, head classes continue to dominate this distribution,  leading to an imbalance rate increase of $\frac{M_1N_C-N_1M_C}{(N_C+M_{C})N_C}, (M_1\gg N_1, N_C/M_C\approx  o (1) )$ and further amplifying model preferences.
    \item When the class distributions are inconsistent, the predictor incorrectly predicts unlabeled samples as head classes, even though most of them actually belong to tail classes. 
     This biased pseudo-labeled distribution causes the model to remain dominated by the head classes, limiting tail class performance.
    
\end{itemize}

\subsubsection{Majority Overconfidence}
Indeed,  according to Fig. \ref{fig3} (Top), we can observe that LFT can obtain relatively precise pseudo-labeled distributions, so why does it still suffer from the “minority bottleneck" problem?
To investigate this issue, we statistically analyze the confidence level of false pseudo-labeling and the total number of them on CIFAR10-LT.
According to the statistics in Fig. \ref{fig3} (bottom), we conclude the following findings.

\begin{itemize}
    \item LP generates numerous false pseudo-labels with low confidence due to its tendency to \textit{under-learn} the training samples by training only the classifier. 
    This leads not only to the model being dominated by easily learnable labeled samples but also to the majority of unlabeled samples being filtered by the masker, further hindering the improvement of tail class performance.
    \item Compared to LP, although LTF significantly reduces the number of erroneous pseudo-labels, it inappropriately increases the prediction confidence level of these pseudo-labels. We term the overconfidence behavior as “\textit{majority overconfidence}".  
\end{itemize}

The primary reason behind this overconfidence is influenced by the inherent bias of the foundation model and an imbalanced training set, LTP produces \textit{biased fitting} for fine-tuning the model to suit downstream data dominated by the head classes.
This behavior results in spurious head class samples with high confidence levels participating in training, which further increases the classification margin of head classes and limits the further performance improvement of tail classes.
\textit{Further analyses are available at Sec. \ref{further}.}

\section{Unbiased Lightweight Fine-tuning}
To effectively mitigate the “minority bottleneck" and “majority overconfidence" problems and achieve unbiased pseudo-labels and classifiers, we present an Unbiased Lightweight Fine-tuning, which consists of two core components: prototype adaptive fitting and dual logit fusion.

\subsection{Prototype Adaptive Fitting}

To reduce the number of false pseudo-labels and their confidence level, we propose a novel Prototype Adaptive Fitting (PAF). Unlike traditional visual models that are overconfident for samples containing certain feature patterns, LFT is overconfident for certain head classes due to biased fitting of downstream tasks. 
In response, we propose confidence-aware adaptive fitting strategy to assist textual prototypes suitable for downstream classification tasks and propose orthogonal loss to refine visual and textual prototypes simultaneously.

Specifically, we obtain “anchor text feature” by feeding the “anchor text” $t$ generated from a template into the pre-trained CLIP model, where the template format is “a photo of a \{\textit{label}\} ”, \{\textit{label}\} is replaced by the category name. Formalized as,

\begin{eqnarray}
 \textbf{C}_{t} = \left \{ \mathbf{c}_{t}^{k} \right \} _{k=1}^{C}, \mathbf{c}_{t}^{k} = \mathcal{T}_{enc}(t) ,
\label{eq2}
\end{eqnarray}
where $\mathbf{c}_{t}^{k}$ denotes the textual prototype of the \textit{k}-th class and $\mathcal{T}_{enc}$ represents the language encoder of pre-training the CLIP model.
Define $\textbf{C}_{v}=\left \{ \mathbf{c}_{v}^{k} \right \}_{k=1}^{C}$ as the visual prototype matrix with Exponential Moving Average (EMA) obtained from the intra-class feature means of the labeled samples in the current batch. 
To motivate the textual prototype to be suitable for long-tailed semi-supervised samples, we introduce the confidence-aware coefficient momentum update $\textbf{C}_{t}$ according to $\textbf{C}_{v}$ as follows,
\begin{eqnarray}
\textbf{c}_t^{k}=(1-\alpha_k) \textbf{c}_t^{k}+\alpha_k \textbf{c}_v^{k},
\label{mt}
\end{eqnarray}
where $\alpha_k$ is obtained from the pseudo-labeled predictive distribution of class \textit{k}, \textit{i.e.}, $\alpha_k =\mu \cdot  \frac{P_u^{k}}{max\{P_u^{i}\}_{1}^{C} } $. $P_u^{i}$ denotes the pseudo-labeled predictive distribution for class \textit{i}, $\mu$ is a weighting. 
By adjusting $\alpha$, we enable the text prototypes to adaptively fit the training data according to the confidence of the pseudo-label distributions. 
When the pseudo-label distribution of a class is small, the model slows down the update of the corresponding prototype, which not only motivates the textual prototype to be fitted reliably but also makes the model more attentive to classes with slow learning rates.

In addition, the obtained textual and visual prototypes may be biased due to the inherent bias of the foundation model and imbalanced training data \cite{liimproving}.
To attain textual and visual prototypes that are uniformly distributed in the hypersphere, we propose a new orthogonal loss that
\begin{eqnarray}
\mathcal{L}_o=\mathcal{H}_{mse}\left ( Sim( \textbf{C}_v, \textbf{C}_{v}^{T}), \textbf{E} \right ) ,
\label{orthogonal}
\end{eqnarray}
where $\mathcal{H}_{mse}$ represents the Mean Square Error loss, $Sim\left ( \cdot  \right ) $ represents the similarity matrix, and $\textbf{E}$ represents the unit matrix consistent with the $Sim\left ( \cdot  \right ) $ dimension. 
Eq. \ref{orthogonal} regulates visual prototypes to be orthogonal to each other by inheriting the maximized visual and textual feature similarity in CLIP (Sec. \ref{similarity}), and asymptotically mitigates the overconfidence problem of textual prototypes to the head classes by Eq. \ref{mt}.

\begin{table*}[t]
\caption{Comparison of top-1 test accuracy $(\%)$ on CIFAR10/100-LT  with setting $\gamma_l = \gamma_u$. We use \textbf{bold} to mark the best results, and \underline{underline} the sub-optimal structures. The subsequent representations are consistent with this.}
  \centering
    \begin{tabular}{l|cc|cc|cc|cc}
    \toprule
    \multirow{4}[6]{*}{Algorithm} & \multicolumn{4}{c|}{CIFAR10-LT} & \multicolumn{4}{c}{CIFAR100-LT} \\
\cmidrule{2-9}          & \multicolumn{2}{c|}{$\gamma=\gamma_{l}=\gamma_{u}=$100} & \multicolumn{2}{c|}{$\gamma=\gamma_{l}=\gamma_{u}=$150} & \multicolumn{2}{c|}{$\gamma=\gamma_{l}=\gamma_{u}=$10} & \multicolumn{2}{c}{$\gamma=\gamma_{l}=\gamma_{u}=$20} \\
\cmidrule{2-9}          & $N_1$=500 & $N_1$=1500 & $N_1$=500 & $N_1$=1500 & $N_1$=50 & $N_1$=150 & $N_1$=50 & $N_1$=150 \\
          & $M_1$=4000 & $M_1$=3000 & $M_1$=4000 & $M_1$=3000 & $M_1$=400 & $M_1$=300 & $M_1$=400 & $M_1$=300 \\
    \midrule
    Supervised & 47.3\text{\scriptsize$\pm$0.95} & 61.9\text{\scriptsize$\pm$0.41} & 44.2\text{\scriptsize$\pm$0.33} & 58.2\text{\scriptsize$\pm$0.29} & 29.6\text{\scriptsize$\pm$0.57} & 46.9\text{\scriptsize$\pm$0.22} & 25.1\text{\scriptsize$\pm$1.14} & 41.2\text{\scriptsize$\pm$0.15} \\
    w/LA  & 53.3\text{\scriptsize$\pm$0.44} & 70.6\text{\scriptsize$\pm$0.21} & 49.5\text{\scriptsize$\pm$0.40} & 67.1\text{\scriptsize$\pm$0.78} & 30.2\text{\scriptsize$\pm$0.44} & 48.7\text{\scriptsize$\pm$0.89} & 26.5\text{\scriptsize$\pm$1.31} & 44.1\text{\scriptsize$\pm$0.42} \\
    \midrule
    FixMatch \cite{sohn2020fixmatch} & 67.8\text{\scriptsize$\pm$1.13} & 77.5\text{\scriptsize$\pm$1.32} & 62.9\text{\scriptsize$\pm$0.36} & 72.4\text{\scriptsize$\pm$1.03} & 45.2\text{\scriptsize$\pm$0.55} & 56.5\text{\scriptsize$\pm$0.06} & 40.0\text{\scriptsize$\pm$0.96} & 50.7\text{\scriptsize$\pm$0.25} \\
    w/DARP \cite{kim2020distribution} & 74.5\text{\scriptsize$\pm$0.78} & 77.8\text{\scriptsize$\pm$0.63} & 67.2\text{\scriptsize$\pm$0.32} & 73.6\text{\scriptsize$\pm$0.73} & 49.1\text{\scriptsize$\pm$0.20} & 58.1\text{\scriptsize$\pm$0.44} & 43.4\text{\scriptsize$\pm$0.87} & 52.2\text{\scriptsize$\pm$0.66} \\
    w/CReST+ \cite{wei2021crest} & 76.3\text{\scriptsize$\pm$0.86} & 78.1\text{\scriptsize$\pm$0.42} & 67.5\text{\scriptsize$\pm$0.45} & 73.7\text{\scriptsize$\pm$0.34} & 44.0\text{\scriptsize$\pm$0.21} & 57.1\text{\scriptsize$\pm$0.55} & 40.6\text{\scriptsize$\pm$0.55} & 52.3\text{\scriptsize$\pm$0.20} \\
    w/ABC \cite{lee2021abc} & 78.9\text{\scriptsize$\pm$0.82} & 83.8\text{\scriptsize$\pm$0.36} & 66.5\text{\scriptsize$\pm$0.78} & 80.1\text{\scriptsize$\pm$0.45} & 47.5\text{\scriptsize$\pm$0.18} & 59.1\text{\scriptsize$\pm$0.21} & 41.6\text{\scriptsize$\pm$0.83} & 53.7\text{\scriptsize$\pm$0.55} \\
    w/DASO \cite{oh2022daso} & 76.0\text{\scriptsize$\pm$0.37} & 79.1\text{\scriptsize$\pm$0.75} & 70.1\text{\scriptsize$\pm$0.63} & 75.1\text{\scriptsize$\pm$0.77} & 50.7\text{\scriptsize$\pm$0.51} & 60.6\text{\scriptsize$\pm$0.71} & 44.1\text{\scriptsize$\pm$0.61} & 55.1\text{\scriptsize$\pm$0.72} \\
    w/L2AC \cite{wang2022imbalanced} & 76.1\text{\scriptsize$\pm$0.45} & 82.1\text{\scriptsize$\pm$0.57} & 70.2\text{\scriptsize$\pm$0.63} & 77.6\text{\scriptsize$\pm$0.53} & -     & 57.8\text{\scriptsize$\pm$0.19} & -     & 52.6\text{\scriptsize$\pm$0.13} \\
    w/ACR \cite{wei2023towards} & 81.6\text{\scriptsize$\pm$0.19} & 84.1\text{\scriptsize$\pm$0.39} & 77.0\text{\scriptsize$\pm$1.19} & 80.9\text{\scriptsize$\pm$0.22} & 51.1\text{\scriptsize$\pm$0.32} & 61.0\text{\scriptsize$\pm$0.41} & 44.3\text{\scriptsize$\pm$0.21} & 55.2\text{\scriptsize$\pm$0.28} \\
    w/BEM \cite{zheng2024bem} & 78.6\text{\scriptsize$\pm$0.97} & 83.0\text{\scriptsize$\pm$0.13} & 72.5\text{\scriptsize$\pm$1.13} & 80.8\text{\scriptsize$\pm$0.67} & 51.3\text{\scriptsize$\pm$0.26} & 61.9\text{\scriptsize$\pm$0.57} & 44.8\text{\scriptsize$\pm$0.21} & 56.1\text{\scriptsize$\pm$0.54} \\
    w/TCBC \cite{li2024twice} & 80.3\text{\scriptsize$\pm$0.45} & 84.0\text{\scriptsize$\pm$0.55} & 75.2\text{\scriptsize$\pm$0.32} & 80.4\text{\scriptsize$\pm$0.58} & -     & 59.4\text{\scriptsize$\pm$0.28} & -     & 53.9\text{\scriptsize$\pm$0.72} \\
    w/CPE \cite{ma2024three} & 80.7\text{\scriptsize$\pm$0.96} & 84.4\text{\scriptsize$\pm$0.29} & 76.8\text{\scriptsize$\pm$0.53} & 82.3\text{\scriptsize$\pm$0.34} & 50.3\text{\scriptsize$\pm$0.34} & 59.8\text{\scriptsize$\pm$0.16} & 43.8\text{\scriptsize$\pm$0.28} & 55.6\text{\scriptsize$\pm$0.15} \\
    w/CCL \cite{zhoucontinuous} & \underline{84.5}\text{\scriptsize$\pm$0.38} & \underline{86.2}\text{\scriptsize$\pm$0.35} & \underline{81.5}\text{\scriptsize$\pm$0.99} & \underline{84.0}\text{\scriptsize$\pm$0.21} & \underline{53.5}\text{\scriptsize$\pm$0.49} & \underline{63.5}\text{\scriptsize$\pm$0.39} & \underline{46.8}\text{\scriptsize$\pm$0.45} & \underline{57.5}\text{\scriptsize$\pm$0.16} \\
    \midrule
    w/LP (Ours)  & 81.2\text{\scriptsize$\pm$0.87} & 84.2\text{\scriptsize$\pm$0.61} & 78.9\text{\scriptsize$\pm$0.94} & 81.0\text{\scriptsize$\pm$0.46} & 68.7\text{\scriptsize$\pm$0.68} & 72.1\text{\scriptsize$\pm$0.53} & 62.1\text{\scriptsize$\pm$0.41} & 68.5\text{\scriptsize$\pm$0.58} \\
    w/LFT (Ours) & 93.2\text{\scriptsize$\pm$0.47} & 95.1\text{\scriptsize$\pm$0.42} & 90.8\text{\scriptsize$\pm$0.51} & 93.6\text{\scriptsize$\pm$0.41} & 78.8\text{\scriptsize$\pm$0.52} & 81.3\text{\scriptsize$\pm$0.51} & 71.2\text{\scriptsize$\pm$0.64} & 77.5\text{\scriptsize$\pm$0.41} \\
    w/ULFine (Ours) & \textbf{96.5}\text{\scriptsize$\pm$0.11} & \textbf{96.7}\text{\scriptsize$\pm$0.07} & \textbf{96.0}\text{\scriptsize$\pm$0.13} & \textbf{96.7}\text{\scriptsize$\pm$0.18} & \textbf{82.1}\text{\scriptsize$\pm$0.27} & \textbf{84.2}\text{\scriptsize$\pm$0.31} & \textbf{79.8}\text{\scriptsize$\pm$0.40} & \textbf{82.3}\text{\scriptsize$\pm$0.18} \\
    \bottomrule
    \end{tabular}%
  \label{table1}%
\end{table*}%

\subsection{Dual Logit Fusion}
Inspired by the complementary nature of the pseudo-labels obtained by the similarity classifier and the linear classifier in \cite{oh2022daso}, we propose Dual Logit Fusion (DLF) to seamlessly align and fuse logits from unbiased textual prototypes and linear probing, respectively.  We expect to obtain unbiased pseudo-labels and classifiers with comprehensive knowledge of logits to further alleviate the “minority bottleneck" problem. \textit{Additional analyses are provided in Sec. \ref{eta-per}}

Specifically, we obtain logit output at the semantic level with the help of the generated unbiased textual prototypes $\textbf{M}_t$ as semantic similarity classifiers, formally as, 
\begin{eqnarray}
\textbf{p}_i^{t}=sim(\textbf{z}_i^{w},\textbf{C}_t)/T,\nonumber
\end{eqnarray}
where $\textbf{z}_i^{w}$ is a visual feature corresponding to the weakly augmented branch and $T$ is a temperature hyperparameter. We define $\textbf{p}_i^{v}$ to represent the logit feature obtained with linear probing.
To eliminate the gap between textual and visual logit features, we seamlessly align them according to their corresponding difference ratios $\beta =\frac{max(\textbf{p}_i^v)-min(\textbf{p}_i^v)}{max(\textbf{p}_i^t)-min(\textbf{p}_i^t)} $. We perform the following conversion of $\textbf{p}_i^t$ according to $\beta$,
\begin{eqnarray}
\hat{\textbf{p}}_i^t=\beta \ast \left ( \textbf{p}_i^t -min(\textbf{p}_i^t)\right ) +min\left ( \textbf{p}_i^v \right ). 
\end{eqnarray}
Then, we fuse the two aligned types of logits,
\begin{eqnarray}
\textbf{p}_i=\eta\textbf{p}_i^v+(1-\eta)\hat{\textbf{p}}_i^t.
\end{eqnarray}
where $\eta$ is a hyperparameter  (fixed at 0.7).
We generate pseudo-label $\tilde{q}_i$ for consistency loss based on $\textbf{p}_i$,
\begin{eqnarray}
\tilde{q}_i =argmax_k(Softmax(\textbf{p}_{i})_k). \nonumber
\end{eqnarray}

Thus, Eq. \ref{eq1} can be rewritten as follows,
\begin{eqnarray}
\begin{aligned}
\tilde{\mathcal{L}}_{F} & =\frac{1}{B_l} \sum_{i=1}^{B_l} \mathcal{H}\left(y_i,p(y|\mathbf{x}_i),\textbf{P}_l\right) \\
& +\frac{1}{B_u} \sum_{j=1}^{B_u} \mathcal{M} \cdot \mathcal{H}\left( \tilde{q}_j,p(y|\mathcal{A}_s({\mathbf{u}}_j) \right),
\label{eq1_t}
\end{aligned}
\end{eqnarray}
where $\textbf{P}_l$ is the class prior distribution of labeled samples for post-hoc logit adjustment\cite{menon2020long}. To sum up, the total loss of our ULFine can be expressed as, 
\begin{eqnarray}
\mathcal{L}_{ULFine}=\tilde{\mathcal{L}}_{F}+\mathcal{L}_o .
\end{eqnarray}
To maintain the inference efficiency and further refine the pseudo-labels, we only retain DLF in the testing phase. Based on Figures 2 and 3, it can be clearly observed that our ULFine can obtain more balanced classification accuracies, more accurate pseudo-labelling predictions as well as fewer false pseudo-labels with lower confidence.

\begin{table*}[t]
    \caption{Comparison of top-1 test accuracy $(\%)$ on CIFAR10-LT and STL10-LT with $\gamma_l \ne \gamma_u$ setting, where $\gamma_l$ is fixed at 100 for CIFAR10-LT and N/A indicates that the data distribution is unknown.}
  \centering
    \begin{tabular}{l|cc|cc|cc|cc}
    \toprule
    \multirow{4}[6]{*}{Algorith} & \multicolumn{4}{c|}{CIFAR10-LT ($\gamma_l \ne \gamma_u$)} & \multicolumn{4}{c}{STL10-LT ($\gamma_u=$N/A)} \\
\cmidrule{2-9}          & \multicolumn{2}{c|}{$\gamma_u=1$ } & \multicolumn{2}{c|}{$\gamma_u=1/100$} & \multicolumn{2}{c|}{$\gamma_l=$ 10} & \multicolumn{2}{c}{$\gamma_l=$20} \\
\cmidrule{2-9}          & $N_1=$500 & $N_1=$1500 & $N_1=$500 & $N_1=$1500 & $N_1=$150 & $N_1=$450 & $N_1=$150 & $N_1=$450 \\
          & $M_1=$4000 & $M_1=$3000 & $M_1=$4000 & $M_1=$3000 & $M=$100k & $M=$100k & $M=$100k & $M=$100k \\
    \midrule
    FixMatch \cite{sohn2020fixmatch} & 73.0\text{\scriptsize$\pm$3.81} & 81.5\text{\scriptsize$\pm$1.15} & 62.5\text{\scriptsize$\pm$0.94} & 71.7\text{\scriptsize$\pm$1.70} & 56.1\text{\scriptsize$\pm$2.32} & 72.4\text{\scriptsize$\pm$0.71} & 47.6\text{\scriptsize$\pm$4.87} & 64.0\text{\scriptsize$\pm$2.27} \\
    w/DARP \cite{kim2020distribution} & 82.5\text{\scriptsize$\pm$0.75} & 84.6\text{\scriptsize$\pm$0.34} & 70.1\text{\scriptsize$\pm$0.22} & 80.0\text{\scriptsize$\pm$0.93} & 66.9\text{\scriptsize$\pm$1.66} & 75.6\text{\scriptsize$\pm$0.45} & 59.9\text{\scriptsize$\pm$2.17} & 72.3\text{\scriptsize$\pm$0.60} \\
    w/CReST \cite{wei2021crest} & 83.2\text{\scriptsize$\pm$1.67} & 87.1\text{\scriptsize$\pm$0.28} & 70.7\text{\scriptsize$\pm$2.02} & 80.8\text{\scriptsize$\pm$0.39} & 61.7\text{\scriptsize$\pm$2.51} & 71.6\text{\scriptsize$\pm$1.17} & 57.1\text{\scriptsize$\pm$3.67} & 68.6\text{\scriptsize$\pm$0.88} \\
    w/CReST+ \cite{wei2021crest} & 82.2\text{\scriptsize$\pm$1.53} & 86.4\text{\scriptsize$\pm$0.42} & 62.9\text{\scriptsize$\pm$1.39} & 72.9\text{\scriptsize$\pm$2.0} & 61.2\text{\scriptsize$\pm$1.27} & 71.5\text{\scriptsize$\pm$0.96} & 56.0\text{\scriptsize$\pm$3.19} & 68.5\text{\scriptsize$\pm$1.88} \\
    w/DASO \cite{oh2022daso}  & 86.6\text{\scriptsize$\pm$0.84} & 88.8\text{\scriptsize$\pm$0.59} & 71.0\text{\scriptsize$\pm$0.95} & 80.3\text{\scriptsize$\pm$0.65} & 70.0\text{\scriptsize$\pm$1.19} & 78.4\text{\scriptsize$\pm$0.80} & 65.7\text{\scriptsize$\pm$1.78} & 75.3\text{\scriptsize$\pm$0.44} \\
    w/ACR \cite{wei2023towards} & 92.1\text{\scriptsize$\pm$0.18} & 93.5\text{\scriptsize$\pm$0.11} & 85.0\text{\scriptsize$\pm$0.99} & 89.5\text{\scriptsize$\pm$0.17} & 77.1\text{\scriptsize$\pm$0.24} & 83.0\text{\scriptsize$\pm$0.32} & 75.1\text{\scriptsize$\pm$0.70} & 81.5\text{\scriptsize$\pm$0.25} \\
    w/BEM \cite{zheng2024bem} & 86.8\text{\scriptsize$\pm$0.47} & 89.1\text{\scriptsize$\pm$0.75} & 70.0\text{\scriptsize$\pm$1.72} & 79.1\text{\scriptsize$\pm$0.77} & 68.3\text{\scriptsize$\pm$1.15} & 81.2\text{\scriptsize$\pm$1.42} & 61.6\text{\scriptsize$\pm$0.98} & 76.0\text{\scriptsize$\pm$1.51} \\
    w/CPE \cite{ma2024three} & 92.3\text{\scriptsize$\pm$0.17} & 93.3\text{\scriptsize$\pm$0.21} & 84.8\text{\scriptsize$\pm$0.88} & 89.3\text{\scriptsize$\pm$0.11} & 73.1\text{\scriptsize$\pm$0.47} & 83.3\text{\scriptsize$\pm$0.14} & 69.6\text{\scriptsize$\pm$0.20} & 81.7\text{\scriptsize$\pm$0.34} \\
    w/CCL \cite{zhoucontinuous} & \underline{93.1}\text{\scriptsize$\pm$0.21} & \underline{93.9}\text{\scriptsize$\pm$0.12} & \underline{85.0}\text{\scriptsize$\pm$0.70} & \underline{89.9}\text{\scriptsize$\pm$0.31} & \underline{79.1}\text{\scriptsize$\pm$0.43} & \underline{84.8}\text{\scriptsize$\pm$0.15} & \underline{77.1}\text{\scriptsize$\pm$0.33} & \underline{83.1}\text{\scriptsize$\pm$0.18} \\
    \midrule
    w/ULFine (Ours) & \textbf{97.6}\text{\scriptsize$\pm$0.08} & \textbf{97.7}\text{\scriptsize$\pm$0.11} & \textbf{96.5}\text{\scriptsize$\pm$0.13} & \textbf{96.9}\text{\scriptsize$\pm$0.12} & \textbf{98.7}\text{\scriptsize$\pm$0.07} & \textbf{99.0}\text{\scriptsize$\pm$0.02} & \textbf{98.7}\text{\scriptsize$\pm$0.08} & \textbf{98.9}\text{\scriptsize$\pm$0.08} \\
    \bottomrule
    \end{tabular}%
  \label{table2}%
\end{table*}%

\section{Experiments}
In this section, we present the main evaluation results of our ULFine on various LTSSL benchmarks. Then, we perform ablation studies on ULFine and provide relevant visualization results. 

\section{Relevant details about datasets and experimental setup}
In this section, we provide an introduction to the relevant datasets and experimental setup.

\subsection{Datasets} We evaluated our method on four benchmark datasets including CIFAR10-TL, CIFAR100-LT, STL10-LT and ImageNet127.
\begin{itemize}
    \item \textbf{CIFAR10/100-TL\cite{krizhevsky2009learning}:} The original CIFAR10/100 dataset contains 10/100 classes with 5000/500 samples per class at $32 \times 32$ resolution. Following previous work \cite{oh2022daso,wei2023towards}, we sample training samples from the dataset to create the imbalanced version. 
    Specifically, for CIFAR10-LT, we evaluated our method in the $N_1=1500, M_1=3000$ and $N_1=500,M_1=4000$ settings. We set the imbalance rates to $\gamma_l=\gamma_u=100$ and $\gamma_l=\gamma_u=150$. We fix $\gamma_l$ and $\gamma_u \in \{1, 1/100\}$ for the uniform and reversed cases. For CIFAR100-LT, we evaluate our method in the $N_1=50,M_1=400$ and $N_1=150, M_1=300$ setting. We set the imbalance rates to $\gamma_l=\gamma_u=10$ and $\gamma_l=\gamma_u=20$. 
    \item \textbf{STL10-LT\cite{coates2011analysis}:} The original STL10 contains 5000 class-balanced labeled samples and 1000K unlabeled samples with unknown distributions. All images are $96 \times 96$ in size. For constructing STL10-LT, we control the imbalance rate of labeled samples to perform sample sampling. We set the imbalance rate $r_l \in \{10, 20\}$ following \cite{oh2022daso,wei2023towards}.
    \item \textbf{ImageNet-127\cite{fan2022cossl}:}  ImageNet-127 is naturally an imbalanced dataset and thus does not require further processing. Moreover, its test set is also imbalanced. In order to save computational resources, following existing methods \cite{wei2023towards,fan2022cossl}, we downsample all images to 32 × 32 or 64 × 64 size.
\end{itemize}

\subsection{Experimental details} 
We perform our experiments on Ubuntu 20.04 OS with inbuilt NVIDIA 3090 GPUs using PyTorch 1.8.0 \cite{paszke2019pytorch}.
Following the previous training regimes \cite{ganerasing}, we use AdaptFormer \cite{chen2022adaptformer} by default to fine-tune the CLIP model due to its effectiveness and efficiency. 
We set the number of iterations to $15k$ with a batch size set to 32 and evaluated every $500$ iterations. We use a standard SGD with a learning rate of 0.03, weight decay set to 5$\times 10^{-4}$, and a momentum factor of 0.9.

\subsection{Main Results}
\paragraph{Baselines} 
Our primary experiments are conducted on four typical benchmark datasets characterized by varying imbalance ratios. For supervised learning,  we train the network using cross-entropy loss using only labeled samples. We compare our method against several competitive LTSSL methods in recent years. These baseline methods include DARP\cite{kim2020distribution}, CReST+\cite{wei2021crest}, ABC\cite{lee2021abc}, DASO \cite{oh2022daso}, L2AC\cite{wang2022imbalanced}, ACR\cite{wei2023towards}, BEM\cite{zheng2024bem}, TCBC\cite{li2024twice}, CPE \cite{ma2024three}, and CCL \cite{zhoucontinuous}. 
For a fair comparison, we follow \cite{zhoucontinuous}, using the same dataset division.

\begin{figure*}[t]
  \centering
   \includegraphics[width=0.98\linewidth]{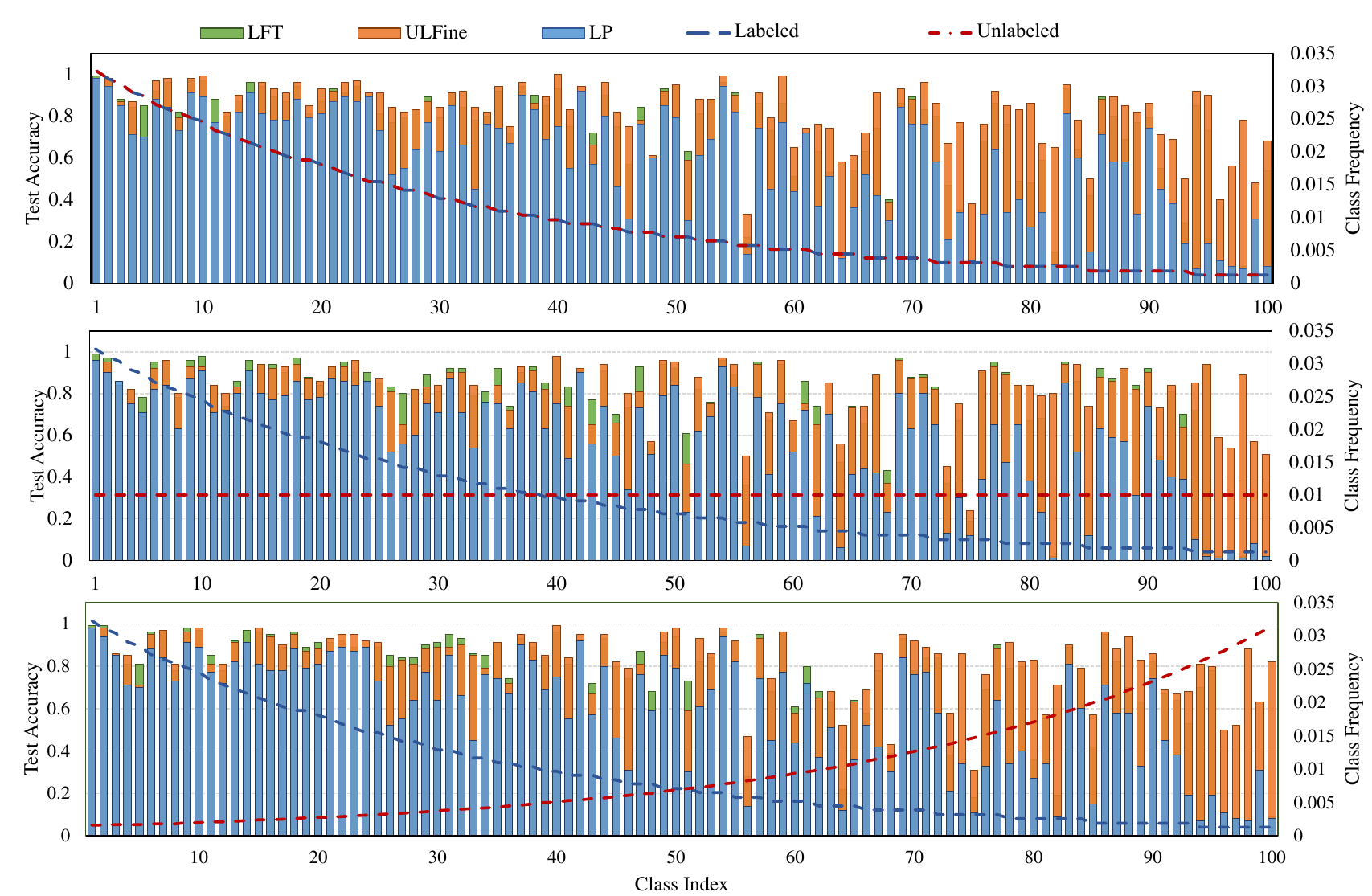}
   \caption{Top-1 classification accuracy per class under different distribution settings of CIFAR100-LT, comparing various strategies.
   }
   \label{cifar100}
\end{figure*}

\paragraph{Results on CIFAR10/100-LT and STL10-LT}
(1) For the \textit{consistent} ($\gamma_l = \gamma_u$) case, the results are shown in Table \ref{table1}. We can observe that ULFine consistently outperforms all comparison methods by a significant margin.
In particular, comparing the previous state-of-the-art method CCL \cite{zhoucontinuous}, our ULFine's top-1 classification accuracy improves by \textbf{19.6}\% on average. Compared to employing CLIP using LP and FFT, ULFine's top-1 classification accuracy improves by \textbf{14.7}\% and \textbf{4.1}\% on average, respectively. 
(2) For the \textit{inconsistent} ($\gamma_l \ne \gamma_u$) case, we present the results in Tables \ref{table2}. Based on the presented experimental results, ULFine outperforms all compared methods across different datasets and settings by significant margins. Among them, the average accuracy of ULFine improves by \textbf{6.7}\% and \textbf{17.8}\% compared to the sub-optimal CCL on CIFAR10-LT and STL10-LT, respectively.
These results indicate that our ULFine can facilitate the model to obtain better generalization.

\paragraph{Results on ImageNet-127} 
To further verify the validity of ULFine, we conduct experiments on the more challenging ImageNet-127 dataset. According to Table \ref{table3}, one can easily find that ULFine achieves the highest test accuracies at different resolutions. Specifically, ULFine's performance improves by \textbf{8.05}\% compared to the sub-optimal ACR+BEM \cite{zheng2024bem}.

\begin{table}[t]
 \caption{Comparison of test accuracy on ImageNet-127.}
  \centering
    \begin{tabular}{l|cc}
    \toprule
    \multirow{2}[4]{*}{Algorith} & \multicolumn{2}{c}{ImageNet-127} \\
\cmidrule{2-3}     & $32\times 32$ & $64\times 64$ \\
    \midrule
    FixMatch \cite{sohn2020fixmatch} &  29.7  & 42.3 \\
    w/DARP \cite{kim2020distribution} &  30.5  & 42.5 \\
    w/DARP+cRT \cite{kim2020distribution} &  39.7  & 51.0 \\
    w/CReST+ \cite{wei2021crest} &  32.5  & 44.7 \\
    w/CReST++LA \cite{wei2021crest} & 40.9  & 55.9 \\
    w/CoSSL \cite{fan2022cossl} & 43.7  & 53.9 \\
    w/TRAS \cite{wei2024transfer}  & 46.2  & 54.1 \\
    w/ACR \cite{wei2023towards}  & 57.2  & 63.5 \\
    w/BEM \cite{zheng2024bem}  & 53.5  & 58.2 \\
    w/ACR+BEM \cite{zheng2024bem} & \underline{58.0}    & \underline{63.9} \\
    \midrule
    w/ULFine (Ours) & \textbf{64.1}  & \textbf{73.9} \\
    \bottomrule
    \end{tabular}%
  \label{table3}%
\end{table}%

\begin{table}[t]
    \caption{Ablation studies of ULFine components.}
  \centering
  \setlength{\tabcolsep}{4pt} 
    \begin{tabular}{cccc|cc}
    \toprule
    LP    & LFT   & PAF   & DLF   & \multicolumn{1}{l}{CIFAR10-LT} & \multicolumn{1}{l}{CIFAR100-LT} \\
    \midrule
    $\checkmark$     &       &       &       & 81.2  & 62.1 \\
    $\checkmark$      & $\checkmark$      &       &       & 93.2  & 71.2 \\
    $\checkmark$      &       & $\checkmark$      & $\checkmark$      & 93.1 & 71.9 \\
    $\checkmark$      & $\checkmark$      & $\checkmark$      &       & \underline{94.1}  & \underline{73.6} \\
    $\checkmark$     & $\checkmark$     & $\checkmark$      & $\checkmark$      & \textbf{96.5}  & \textbf{79.8} \\
    \bottomrule
    \end{tabular}%
  \label{table4}
\end{table}%

\begin{figure*}[t]
  \centering
   \includegraphics[width=0.9\linewidth]{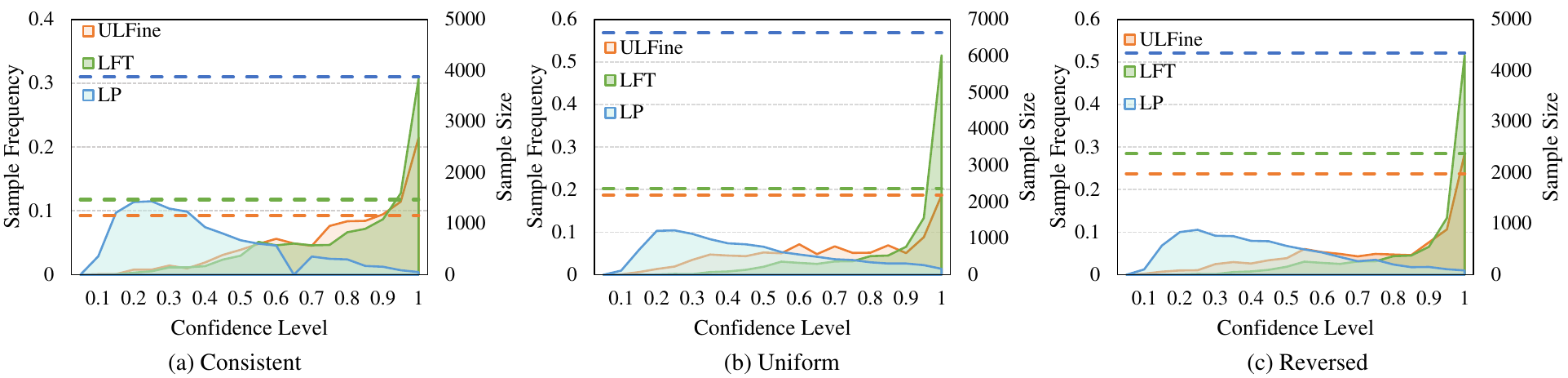}
   \caption{The vertical axes (left and right) indicate the confidence level (area plot) and the sample size (dashed line) of false pseudo-labeling.
   }
   \label{CIFAR100-confidence}
\end{figure*}

\subsection{Further problem analysis}
\label{further}
To further clarify the “majority overconfidence" problem suffered by the Lightweight Fine-Tuning (LFT) model, we conducted experimental analyses on the additional CIFAR100-LT dataset, where labeled and unlabeled samples corresponded to maximum intra-class sample sizes of $N_1$ = 50 and $M_1$ = 400, respectively, and the imbalance rate of the labeled samples corresponded to $\gamma_l$ = 20. In addition, "Consistent", "Uniform", and "Reversed" correspond to the imbalance rate of the unlabeled samples with $\gamma_u$=20, $\gamma_u$=,1 and $\gamma_u$=1/20, respectively. 

As shown in Fig. \ref{CIFAR100-confidence}, across different experimental settings, using Linear Probing (LP) produced a large number of false pseudo-labels. Although employing LTF significantly reduces the number of samples with erroneous pseudo-labeling, it undesirably increases the confidence level of these samples. In the semi-supervised training paradigm, these erroneous samples with high confidence are mistakenly added to the training by the masker as correct samples, exacerbating model bias. In contrast, our ULFine not only further reduces the number of false pseudo-labels but also decreases their confidence level. This validates that ULFine significantly mitigates the “majority overconfidence" problem, and shows that our method can prevent incorrect pseudo-labels from interfering with model training, as well as promote the model to produce unbiased pseudo-labels and classifiers.

\subsection{Experimental results on the balanced dataset.}

\begin{table}[t]
  \centering
  \caption{Experimental results of different algorithms on the balanced dataset CIFAR100.}
    \begin{tabular}{l|ccc}
    \toprule
    \multirow{2}[4]{*}{Algorithm} & \multicolumn{3}{c}{CIFAR100} \\
\cmidrule{2-4}          & N4    & N25   & N100 \\
    \midrule
    FixMatch \cite{sohn2020fixmatch} & 77.10  & 84.05 & 86.17 \\
    DebiasPL\cite{wang2022debiased} & 79.57  & 84.01 & 86.16 \\
    FineSSL \cite{ganerasing} & \textbf{80.44} & \underline{84.51} & \underline{86.66} \\
    \midrule
    ULFine (Ours)  & \underline{80.27} & \textbf{85.07} & \textbf{86.91} \\
    \bottomrule
    \end{tabular}%
  \label{balanced}%
\end{table}%

To validate the generalization capability and effectiveness of ULFine, we conduct experiments on the balanced CIFAR100 dataset, where $N*$ denotes the number of labeled samples per class. As evidenced in Table \ref{balanced}, ULFine exhibits comparable performance to the SOTA FineSSL, which is a foundation model-based (CLIP) approach. In particular, ULFine achieves superior classification accuracy under the N25 and N100 settings, outperforming FineSSL by \textbf{0.56\%} and \textbf{0.25\%}, respectively. The relatively small performance gap observed in the N4 setting can be attributed to ULFine's primary focus on addressing class imbalance rather than few-shot learning. Consequently, the performance of ULFine is more advantageous as the amount of labeled data increases.

\subsection{Ablation Studies and Visualization Results.}

\subsubsection{Ablation studies for different components}
On the CIFAR10/100-LT dataset, we conduct a series of ablation studies on the components included in ULFine, and the relevant experimental results are summarised in Table \ref{table4}.
We can observe that the PAF (Prototype Adaptive Fitting)  and DLF (Dual Logit Fusion) proposed in this paper provide a significant boost to the model. In particular, using only DLF improves the classification accuracy by \textbf{2.4}\% and \textbf{6.2}\% on CIFAR10-LT and CIFAR100-LT, respectively. 
Additionally, we observe that by solely integrating the proposed PAF and DLF methods based on Linear Probing (LP), comparable performance can be achieved, even without employing Lightweight Fine-Tuning (LFT). 
This underscores the potential of our approach to enhance the adaptation of the foundation model to long-tailed semi-supervised data, even when training is confined to classifiers alone.

\subsubsection{Visualization of classification performance}
To validate ULFine's ability to achieve relatively unbiased classification, we conduct a comprehensive visualization study of per-class accuracy across different data distributions on CIFAR100-LT. As shown in Fig. \ref{cifar100}, while LP and LFT achieve superior performance on head classes corresponding to the labeled data, they exhibit significant performance degradation on tail classes,  which is consistent with the "minority bottleneck" phenomenon.  ULFine, in contrast, achieves flatter classification accuracies,  indicating that it can achieve a more balanced classifier. Notably, ULFine substantially improves the classification accuracy of tail classes, thereby further confirming that the proposed approach can effectively mitigate the "minority bottleneck" problem.

\subsubsection{Visualization of similarity matrix}
\label{similarity}
\begin{figure*}[t]
  \centering
   \includegraphics[width=0.9\linewidth]{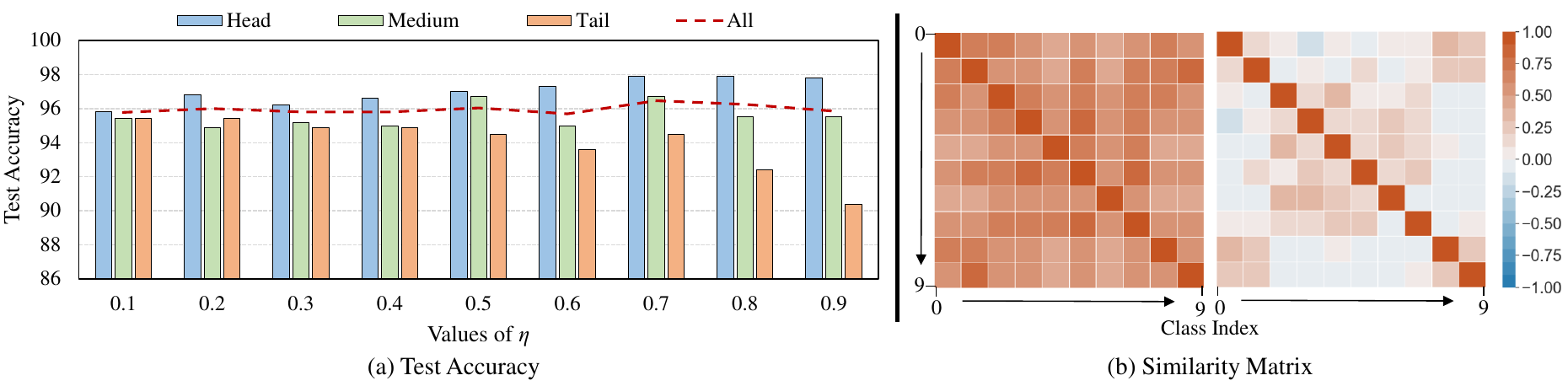}
   \caption{(a) Effect of different a on model performance on the CIFAR10-LT dataset.  (b) Comparison of similarity matrices between textual prototypes before and after using $\mathcal{L}_o$ on the CIFAR10-LT dataset.
   }
   \label{eta}
\end{figure*}

To verify the validity of the proposed orthogonal loss (\textit{i.e.}, Eq. \ref{orthogonal}), we visualise the similarity matrix corresponding to the textual prototypes on the CIFAR10-LT dataset. As shown in Fig. \ref{eta}(b), the left and right represent the corresponding confusion matrices before and after using $\mathcal{L}_o$, respectively. We can observe that the pairwise similarity between different text prototypes decreases significantly after using $\mathcal{L}_o$, implying that it promotes mutual orthogonality between prototypes.

\begin{table}[t]
    \caption{Comparison of different fine-tuning methods. We use \textbf{bold} to mark the best results, and \underline{underline} the sub-optimal structures.}
  \centering
    \begin{tabular}{c|l|c|c}
    \toprule
    \multicolumn{2}{c|}{\multirow{4}[6]{*}{Algorithm}} & \multicolumn{1}{l|}{CIFAR10-LT} & \multicolumn{1}{l}{CIFAR100-LT} \\
\cmidrule{3-4}    \multicolumn{2}{c|}{} & {$\gamma_l=\gamma_u$=100} & {$\gamma_l=\gamma_u$=10} \\
\cmidrule{3-4}    \multicolumn{2}{c|}{} & $N_1$=500 & $N_1$=50 \\
    \multicolumn{2}{c|}{} & $M_1$=4000 & $M_1$=400 \\
    \midrule
    \multicolumn{2}{l|}{CCL \cite{zhoucontinuous}} & 84.50  & 53.50 \\
    \midrule
    \multicolumn{2}{l|}{Linear Probing} & 81.20  & 68.70 \\
    \midrule
    \multirow{7}[2]{*}{ULFine} & BitFit & 90.18 & 78.09 \\
          & VPT-last & 91.73 & 78.51 \\
          & VPT-shallow & 94.23 & 78.96 \\
          & VPT-deep & \underline{96.01} & 81.71 \\
          & Adapter & 95.70  & 81.70 \\
          & LoRA  & 95.98 & \underline{81.84} \\
          & AdaptFormer & \textbf{96.46} & \textbf{82.10} \\
    \bottomrule
    \end{tabular}
  \label{FT}%
\end{table}%

\subsection{Impact of different $\eta$ on performance}
\label{eta-per}

In order to explore the complementary properties of the two types logits in the Dual Logit Fusion component, we observe how the performance of the model changes with the change of the weight coefficient $\eta$ in Eq. 7 over CIFAR10-LT ($N_1$ = 500, $M_1$ = 4000, $\gamma_l$ = 100 and $\gamma_u$ = 100). We counted the overall performance, the head class performance, the medium class performance, and the tail class performance, as shown in Fig. \ref{eta} (a). 

We can observe that the overall performance of the model fluctuates only slightly as $\eta$ changes. This indicates that our model exhibits excellent stability and generalization ability, and therefore can better adapt to complex real-world scenarios. 
Furthermore, it is not difficult to find that as $\eta$ increases, \textit{i.e.}, the logit weights obtained by the linear probing gradually increase, the head class performance gradually improves while the tail class performance shows a decreasing trend, and vice versa. This precisely demonstrates that the logits obtained from semantic prototypes and linear probing are complementary properties, \textit{i.e.}, the logits obtained from linear probing are biased towards the head classes, while the logits obtained from semantic prototypes are biased towards the tail classes. 
These properties are consistent with the observations of \cite{oh2022daso}.
Ultimately, we obtain unbiased logits for semi-supervised imbalance scenarios by seamlessly fusing these two types of logits.

\subsection{Impact of different fine-tuning strategies on performance}
This paper proposes an unbiased lightweight fine-tuning strategy as a general framework that can be applied to different fine-tuning strategies, including but not limited to Bias-terms Fine-tuning (BitFit) \cite{zaken2021bitfit}, Visual Prompt Tuning (VPT) \cite{jia2022visual}, Adapter \cite{houlsby2019parameter}, Low-Rank Adapter (LoRA) \cite{hulora} and AdaptFormer \cite{chen2022adaptformer}. 
To verify the inclusiveness of the methods in this paper, we test ULFine on the CIFAR10/100-LT dataset using seven different lightweight fine-tuning strategies.

The experimental results in Table \ref{FT} show that using arbitrary fine-tuning strategies corresponds to performance that significantly outperforms both the state-of-the-art baseline method (CCL) and Linear Probing. Specifically, using AdaptFormer yields optimal performance on both datasets, while using BitFit yields the worst performance compared to the other fine-tuning strategies listed.

\begin{table*}[h]
      \caption{Comparison of different metrics with the baseline methods on the CIFAR100-LT dataset. The "Average Accuracy" indicates the average performance of the model across different imbalance rates with a consistent distribution of labeled and unlabeled datasets.  The subsequent representations are consistent with this.}
  \centering
    \begin{tabular}{l|c|c|c|c|c}
    \toprule
    \multicolumn{1}{l|}{Algorith} & \multicolumn{1}{c|}{Epochs} & \multicolumn{1}{c|}{Backbone}& \multicolumn{1}{c|}{Learnable Params($\approx $)} & \multicolumn{1}{c|}{Batchsize} & \multicolumn{1}{c}{Average Accuracy} \\
    \midrule
    FixMatch \cite{sohn2020fixmatch} & 2.5$\times 10^5$ & Wide ResNet-28-2 & 1.50 M  & 64    & 48.10 \\
    w/DARP \cite{kim2020distribution} & 2.5$\times 10^5$ & Wide ResNet-28-2 & 1.50 M  & 64    & 50.78 \\
    w/CReST+ \cite{wei2021crest} & 2.5$\times 10^5$ & Wide ResNet-28-2 & 1.50 M  & 64    & 48.50 \\
    w/ABC \cite{lee2021abc} & 2.5$\times 10^5$ & Wide ResNet-28-2 & 1.50 M  & 64    & 50.48 \\
    w/DASO \cite{oh2022daso} & 2.5$\times 10^5$ & Wide ResNet-28-2 & 1.50 M  & 64    & 52.28 \\
    w/L2AC \cite{wang2022imbalanced} & 2.5$\times 10^5$ & Wide ResNet-28-2 & 1.50 M  & 64    & - \\
    w/ACR \cite{wei2023towards} & 2.5$\times 10^5$ & Wide ResNet-28-2 & 1.50 M  & 64    & 52.90 \\
    w/BEM \cite{zheng2024bem} & 2.5$\times 10^5$ & Wide ResNet-28-2 & 1.50 M  & 64    & 53.53 \\
    w/TCBC \cite{li2024twice} & 2.5$\times 10^5$ & Wide ResNet-28-2 & 1.50 M  & 64    & - \\
    w/CPE \cite{ma2024three} & 2.5$\times 10^5$ & Wide ResNet-28-2 & 1.50 M  & 64    & 52.38 \\
    w/CCL \cite{zhoucontinuous} & 2.5$\times 10^5$ & Wide ResNet-28-2 & 1.50 M  & 64    & \underline{55.33} \\
    \midrule
    w/ULFine (Ours) & 1.5$\times 10^4$ & ViT-B/16 & 0.10 M & 32    & \textbf{82.10} \\
    \bottomrule
    \end{tabular}%
  \label{detail}%
\end{table*}%

\begin{table}[!t]
  \centering
  \caption{Training time comparison (in seconds) across different methods, where 'FM' indicates whether the foundation model is used or not.}
    \begin{tabular}{l|c|ccc}
    \toprule
    Algorith & FM & Per Step & Steps & Total time \\
    \midrule
    FixMatch \cite{sohn2020fixmatch} & $\times$     & 0.062 & 2.5$\times 10^5$ & 15500 \\
    CPE \cite{ma2024three}  & $\times$     & 0.188 & 2.5$\times 10^5$ & 47000 \\
    \midrule
    FineSSL \cite{ganerasing} & $\checkmark$      & 0.642 & 1.5$\times 10^4$ & 9630 \\
    ULFine (Ours)  & $\checkmark$     & 0.498 & 1.5$\times 10^4$& 7470 \\
    \bottomrule
    \end{tabular}%
  \label{times}%
\end{table}%

\subsection{Comparison of experimental details of different methods}
To verify the efficiency of our method, we compare the experimental details and average accuracies of ULFine with existing methods on the CIFAR100-LT dataset. According to Table \ref{detail}, we can observe that ULFine requires training only 1.5$\times 10^4$ epochs, which reduces the training cost by nearly \textbf{10} times compared to the baseline method's 2.5$\times 10^5$. In addition, ULFine requires significantly fewer learnable parameters and batch sizes, and significantly increases the model's average accuracy. Specifically, the average accuracy of ULFine increased by \textbf{26.77}\% compared to the sub-optimal CCL.
This is because ULFine introduces only a small number of task-specific parameters and inherits the excellent generalization performance of the foundation model, thus ULFine not only exhibits fast convergence but also significantly improves model performance.

To further evaluate ULFine's efficiency, Table \ref{times} compares the training times of different methods under identical experimental setups. The results demonstrate that ULFine significantly reduces total training time compared to foundation-model-free approaches (FixMatch and CPE). Moreover, while handling more challenging LTSSL tasks, ULFine achieves a \textbf{2160s (22\%)} faster training speed than FineSSL (SSL tasks with foundation models). These findings confirm ULFine's consistent training efficiency improvements over both conventional training-from-scratch methods and existing foundation-model-based approaches.

\section{Conclusion}
In this paper, we explore the impact of employing foundation models like CLIP in different ways on long-tailed semi-supervised tasks. We observe that simply employing the existing tuning strategies suffers from the “minority bottleneck" and “majority overconfidence" problems. 
To alleviate these issue, we propose a simple and effective Unbiased Lightweight Fine-tuning strategy, ULFine, which consists of two core components, Prototype Adaptive Fitting and Dual Logit Fusion. ULFine not only exhibits faster convergence but also consistently outperforms the compared baseline methods across multiple benchmark datasets and experimental setups. We hope that our method can provide some meaningful insights to facilitate the further development of long-tailed semi-supervised learning.

\section*{Acknowledgments}
The authors would like to thank the support from the National Natural Science Foundation of China (62376126, 62076124, 62106102), the Natural Science Foundation of Jiangsu Province (BK20210292), the Fundamental Research Funds for the Central Universities (NS2024058), and the Hong Kong Scholars Program (XJ2023035).

\bibliographystyle{IEEEtran}
\bibliography{main}

\newpage

\vfill

\end{document}